\documentclass[letterpaper]{article} 
\usepackage{aaai23}  
\usepackage{times}  
\usepackage{helvet}  
\usepackage{courier}  
\usepackage[hyphens]{url}  
\usepackage{graphicx} 
\usepackage{natbib}  
\usepackage{caption} 
\usepackage{algorithm}
\usepackage{algorithmic}

\usepackage{amssymb}
\usepackage{booktabs}
\usepackage{tikz}
\usepackage{subcaption}
\usepackage{multirow}
\usepackage{newfloat}
\usepackage{listings}
\DeclareCaptionStyle{ruled}{labelfont=normalfont,labelsep=colon,strut=off} 
\floatstyle{ruled}
\newfloat{listing}{tb}{lst}{}
\floatname{listing}{Listing}
\title{ROIFormer: Semantic-Aware Region of Interest Transformer for Efficient Self-Supervised Monocular Depth Estimation}
\author{
    Daitao Xing\textsuperscript{\rm 1}\thanks{Corresponding author.The work was done when Xing was an intern with OPPO},
    Jinglin Shen\textsuperscript{\rm 2},
    Chiuman Ho\textsuperscript{\rm 2},
    Anthony Tzes\textsuperscript{\rm 3}
}
\affiliations{
    \textsuperscript{\rm 1} New York University, USA\\
    \textsuperscript{\rm 2} OPPO US Research Center, USA\\
    \textsuperscript{\rm 3} New York University Abu Dhabi and Center for Artificial Intelligence and Robotics, UAE\\
    daitao.xing@nyu.edu, {jinglin.shen, chiuman.ho}@oppo.com,  anthony.tzes@nyu.edu
}

\usepackage{bibentry}

\begin{document}

\maketitle

\begin{abstract}
The exploration of mutual-benefit cross-domains has shown great potential toward accurate self-supervised depth estimation. In this work, we revisit feature fusion between depth and semantic information and propose an efficient local adaptive attention method for geometric aware representation enhancement. Instead of building global connections or deforming attention across the feature space without restraint, we bound the spatial interaction within a learnable region of interest. In particular, we leverage geometric cues from semantic information to learn local adaptive bounding boxes to guide unsupervised feature aggregation. The local areas preclude most irrelevant reference points from attention space, yielding more selective feature learning and faster convergence. We naturally extend the paradigm into a multi-head and hierarchic way to enable the information distillation in different semantic levels and improve the feature discriminative ability for fine-grained depth estimation. Extensive experiments on the KITTI dataset show that our proposed method establishes a new state-of-the-art in self-supervised monocular depth estimation task, demonstrating the effectiveness of our approach over former Transformer variants.
\end{abstract}

\section{Introduction}
\label{sec:intro}
Accurate depth estimation is critical for many applications in computer vision and robotics fields such as perception, navigation, and path planning. The advancements in deep learning have brought significant breakthroughs in the accuracy of depth estimation methods in recent years. While supervised learning-based methods like  \cite{ranftl2021visionTransdense} achieved remarkable performance on pixel-wise dense predictions of monocular images, the requirement of a large number of dense labels for training and the excessive spending for acquiring those datasets using LiDAR constraint their usage in real-world applications. Instead, self-supervised depth estimation methods, which learn the depth values using only monocular or stereo image sequences, have become more popular.\\
\indent{}Monocular self-supervised depth estimation utilizes the photometric loss and smoothness constraints of consecutive frames in image sequences to simultaneously learn the depth and pose networks. Despite the notable achievement, self-supervised methods, which only rely on similarity constraints, still have large performance gaps with the supervised methods. \cite{lyu2021hrdepth} shows that the bottlenecks come from the inaccurate depth estimation, especially at object boundaries due to the moving objects, ambiguity in low-texture regions, reflective surfaces, occlusion, and the uncertainty of pose estimation. However, the sole consistency constraints of RGB images are insufficient to reduce those effects, which instead require external modalities to provide stronger geometric information.\\
\begin{figure}
\centering
\includegraphics[width=\linewidth]{ 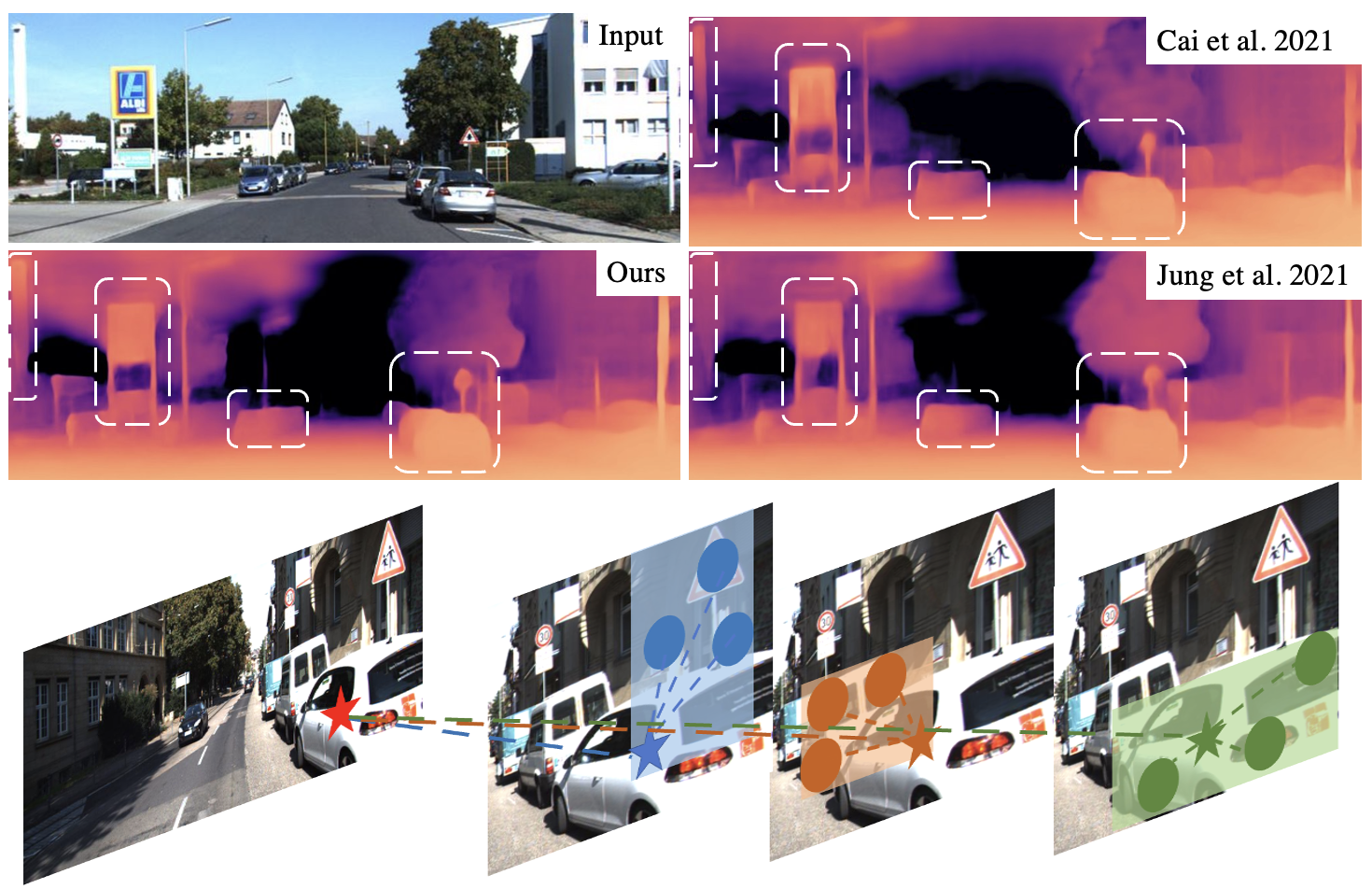}
\caption{Depth prediction (top two rows) from a single image on KITTI with semantic guidance shows geometry consistency preservation in local uncertainty areas. ROI Former overview (bottom row) shows that the depth of the red star requires attention from various semantic regions.}
\label{fig:introduction}

\end{figure}
\indent{}Recent works, instead, leverage semantic information to improve monocular depth predictions via incorporating geometric guidance. While most of those works incorporate the semantics information explicitly, fewer works focus on designing cross-domain feature aggregation strategies to optimize the intermediate depth representations. Recently, Transformer based attention methods demonstrated their superiority over traditional CNNs in many vision tasks. \cite{jung2021fineFSRE} proposes a cross-modality attention module to refine depth features progressively on multi-scales and obtains significant improvement. However, the interaction is restricted within corresponding features to avoid the computational overhead in the classic Transformer methods.
\cite{zhu2020deformable} provides an alternative method to Transformer with significantly reduced parameters and faster convergence speed. However, we found that the performance of deformable attention drops dramatically when high-resolution images are used as inputs. We argue that this is due to the attention module collapse and failure to locate relative information as the feature space becomes too large. \\
\indent{}The aforementioned analysis related to semantics-guided depth estimation and the success of attention-based feature aggregation indicates that these are mainly determined by two aspects: (1) to provide a dynamic attention region assigned to each reference point, which covers the intact local semantic information and excludes irrelevant points regardless of the spatial size of the feature maps, (2) locate the semantic positions within search areas to update the reference point. Subsequently, we propose a Region-of-Interests (ROIs) guided deformable attention module, named ROIFormer, which performs deformable attention within learnable adaptive local region proposals. Inspired by \cite{yang2018metaanchor} and \cite{sun2021sparse}, those object-aware proposals can be inferred directly from semantically feature maps via lightweight networks. Unlike \cite{yang2018metaanchor}, which creates proposals with deformation on meta-anchors, we merge the proposal generation inside multi-head attention modules and define attention areas implicitly according to information from different semantic levels. The deformable attention is then performed within constrained regions to find the most relative semantic features. With the search space restrained into object-aware local areas, the search complexity is dramatically reduced, resulting in robust feature enhancement and fast convergence.\\
\indent{}Despite the depth consistency provision within each connected segment block using the semantic features, the instance level information is still missing, which results in uncertainty on the boundaries. Thus, we consider the spatial relationship between instance objects with the crowded areas (roads, sidewalks, and buildings) in 3D space. The points projected into the 3D space based on the estimation depth and intrinsic camera model should stay close to the nearby points within the same category and keep a reasonable distance to the reference crowded areas. Instead, the points far away from the reference points are masked as outliers which should not be used to calculate photometric similarity loss. \\
Overall, our main contributions are summarized as follows:
  \begin{itemize}
  \item We provide a detailed comparison between different feature fusion strategies for efficient self-supervised depth estimation, indicating that search space complexity is critical for model convergence and performance improvement.
  \item We propose the ROIFormer, which guides the attention in local areas to most relative semantics information in an unsupervised and efficient way.
  \item The suggested self-supervised depth estimation with semantics guidance network achieves state-of-the-art performance on varies settings.  
\end{itemize}
\section{Related Work}
\label{sec:related}
\subsection{Self-Supervised Monocular Depth Estimation}
Significant improvement has been made since \cite{zhou2017unsupervised} proposed the generalized framework, which enables supervision from consecutive frames via ego-motion. Later works focus on a more elegant loss design to filter unreliable propagation. \cite{yang2020d3vo} models the photometric uncertainties of pixels on input images. \cite{shu2020feature} proposed to feature metric loss to stabilize the loss landscape process. \cite{bian2019unsupervised} upgraded the geometry consistency loss for scale-consistent predictions. 
In order to provide more geometry information for self-supervised learning, \cite{ranjan2019competitive,wang2019unos,zhao2020towards,petrovai2022exploiting,vi2020mdanet} integrate optical flows and pseudo labels for extra constraints. \cite{packnet} and \cite{lyu2021hrdepth} proposed optimized architectures for more efficient depth estimation. 
\begin{figure*}[t]
\centering
\includegraphics[width=0.8\textwidth]{ 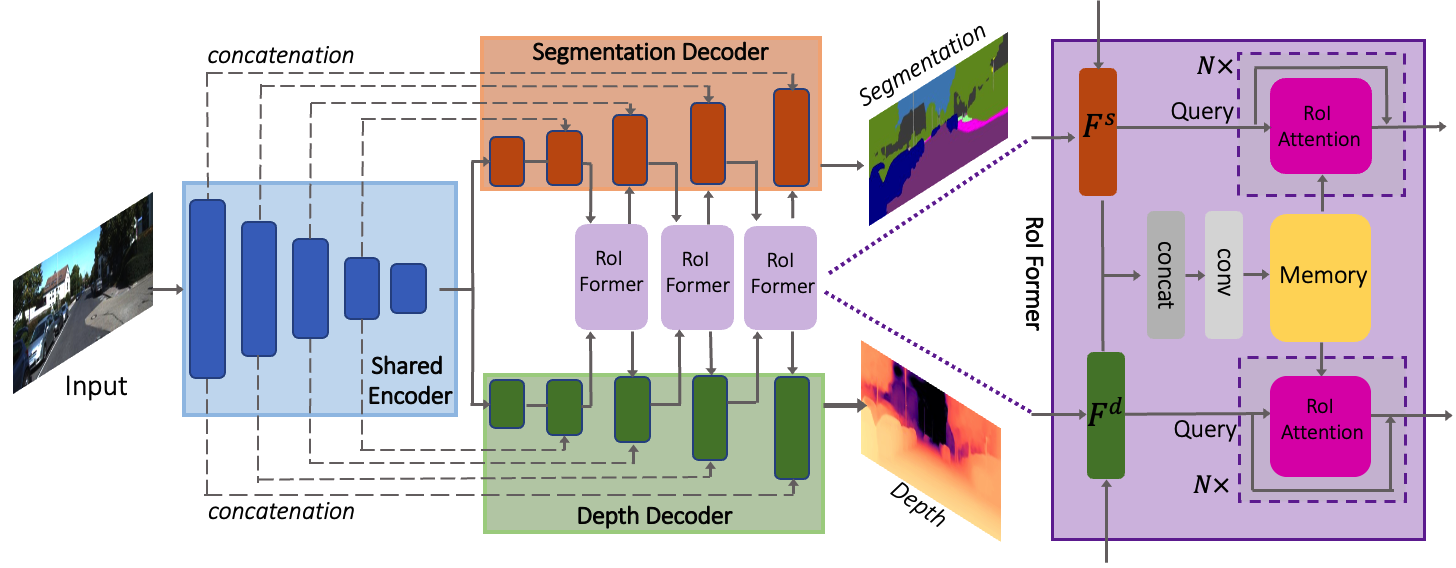}
\caption{Proposed framework for self-supervised monocular depth estimation with semantic guidance. The segmentation and depth branches share the same backbone as the encoder, and capture mutual benefits information from other domains with adaptive local attention modules.}
\label{fig:framwork}
\end{figure*}
\subsection{Semantic-guidance for Depth Estimation}
Semantic segmentation with strong geometry knowledge is widely used for the promotion of depth estimation. \cite{lee2021learning} improves the performance using an instance-aware geometric consistency loss. \cite{zhu2020edge} explicitly measures the border consistency between segmentation and depth and minimizes it in a greedy manner. \cite{casser2019depthStruct2Depth,klingner2020sgdepth} stabilize the photometric loss by removing moving dynamic-class objects. \cite{chen2019towards} performs region-aware depth estimation by enforcing semantics consistency, while \cite{guizilini2020semantically} uses pixel-adaptive convolutions to produce semantic-aware depth features via assigning weights to features within a local window. \cite{li2021learning} and  \cite{jung2021fineFSRE} design a cross-task attention module to refine depth features progressively on multi-scales.  \cite{tosi2020distilled} and \cite{cai2021xdistill} apply knowledge distillation from semantic segmentation to depth estimation with a learnable domain transfer network.
\subsection{Efficient Attention Network}
Transformer~\cite{carion2020enddetr} indicates a stronger performance over traditional CNNs. After that, lot of works, including \cite{dai2017deformable, xia2022vision, xie2020spatially, chen2021dpt, yue2021vision, liu2021swin, wang2021pyramid} design efficient multi-scale attention for detection and classification tasks. \cite{xie2021segformer} and \cite{ranftl2021visionTrans} customized Transformer with encoder and decoder frameworks for dense prediction tasks. More recently, \cite{bae2022monoformer} and \cite{li2022depthformer} migrate Transformer into supervised depth estimation. \cite{johnston2020self, zhao2022monovit} and \cite{jung2021fineFSRE} integrate attention into self-supervised depth estimation and obtain significant improvement. \cite{nguyen2022boxer} samples points over 3X3 transformed grids guided from ground truth boxes for efficient detection. However, those methods rely on either global dependence or local interactions. Instead, we select attention areas adaptively for optimal efficiency.  

%
\section{Proposed Methods}
\label{sec:method}
\subsection{Self-Supervised Depth Estimation}
Self-supervised monocular depth estimation utilizes source images $I_{t-1}$ and $I_{t+1}$ to build reference image $\hat{I}_{t' \rightarrow t}$ for target image $I_t$ via geometric transformation. We learn a scale-ambiguous depth $\hat{D}_t$ map and a corresponding semantic segmentation map $\hat{S}_t$ from the multi-task network. With known intrinsic camera parameter matrix $K \in \mathbb{R}^{3 \times 3}$, the pixels $p \in I_t$ with homogeneous coordinates $u$ are projected to the 3D-space, resulting in a point cloud $P_t$ with pixel value and semantic information. The pose network, instead, outputs the 6 DOF transformation $T_{t \rightarrow t'} | t' \in \{t-1, t+1\}$, rotates, shifts, and re-projects the point cloud to get pixel coordinates in image $I_{t'}$. The correspondence between $I_t$ with $I_t'$ can be summarized as:
\begin{equation}
p^{\prime}=K T_{t \rightarrow t'} \hat{D}_{t} K^{-1} p.
\end{equation}
The reference image $\hat{I}_{t' \rightarrow t}$ for $I_t$ is obtained via the interpolation on $I_t'$ according to the reprojected pixel coordinates $p'$. The objective is to minimize the similarity between $\hat{I}_{t' \rightarrow t}$ and $I_t$ by calculating the Structural Similarity (SSIM) loss and L1 distance, as:
\begin{equation}
\hspace*{-3mm}L_{p,t'}= \alpha \frac{1-\mbox{SSIM}\left(I_{t}, \hat{I}_{t' \rightarrow t}\right)}{2}+(1-\alpha)\left|I_{t}-\hat{I}_{t' \rightarrow t}\right| .
\end{equation}
We incorporate the minimum reprojection error by selecting the per-pixel minimum value cross all similarity losses, where $L_{p}=\min _{t^{\prime}} L_{p, t'}$. Following (Godard et al. 2017), we include an edge-aware term to smooth the depth in low gradient areas, defined as:
\begin{equation}
L_{s}=\left|\partial_{x} D_{t}\right| e^{-\left|\partial_{x} I_{t}\right|}+\left|\partial_{y} D_{t}\right| e^{-\left|\partial_{y} I_{t}\right|}.
\end{equation}
The smoothness loss reinforces the depth similarity between pixels with small differences in grey values.\\
\indent{}It is proven that the multi-task joint training stimulates mutual benefits and results in significant improvements to both tasks. Therefore, we adopt pre-computed segmentation maps from \cite{jung2021fineFSRE} as ground truth and train the semantic segmentation branch with cross entropy loss.
%
\subsection{Feature Enhancement with ROI Attention}
\begin{figure}[t]
\centering
\includegraphics[width=\linewidth]{ 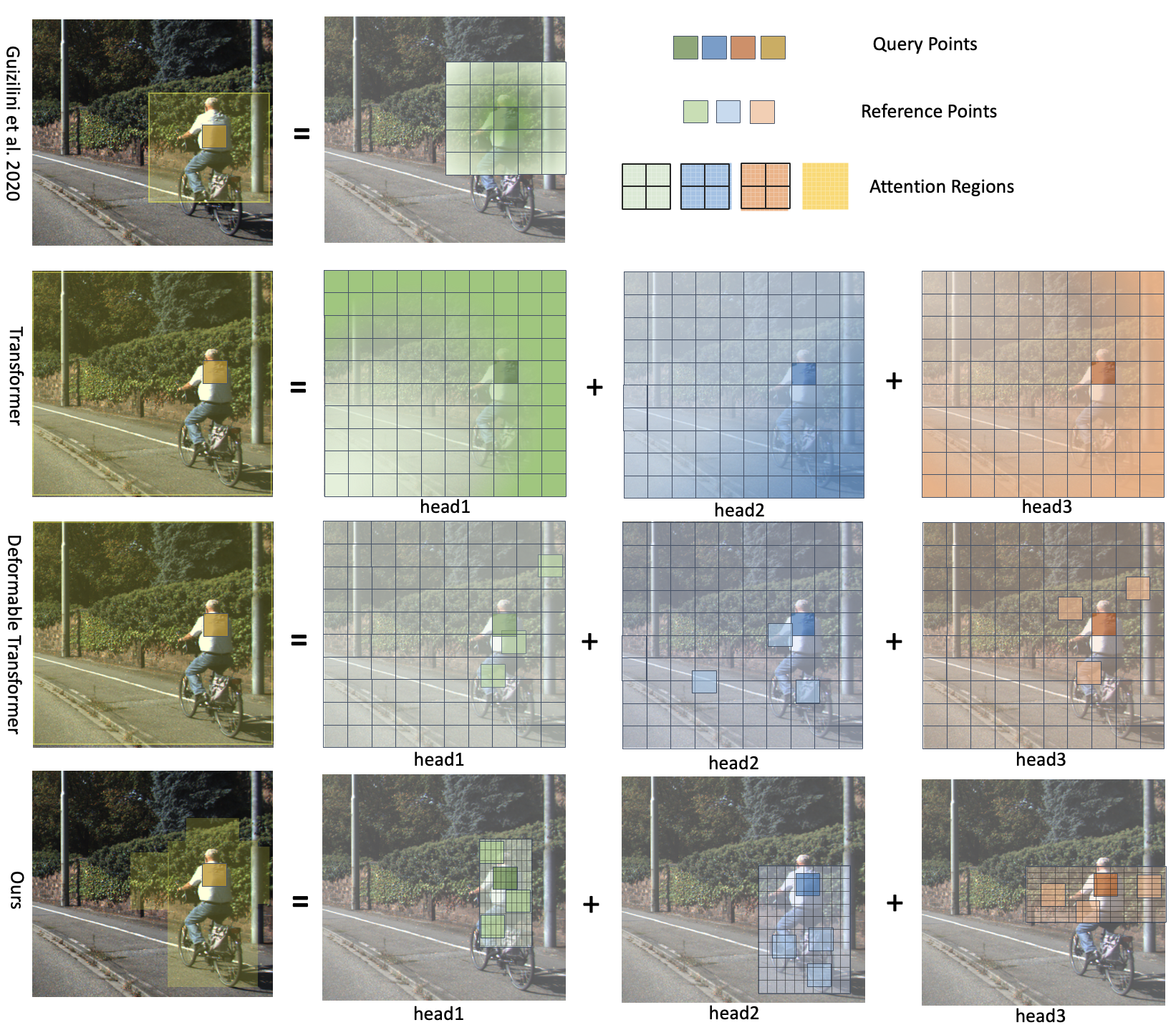}
\caption{Comparison of feature fusion strategies for depth estimation: adaptive convolution (top row), global dense attention (second row), global sparse attention (third row) and our adaptive sparse attention (bottom row).}
\label{fig:atten_compare}
\end{figure}

The principle of jointly training depth estimation and segmentation is to distill the semantic and position information and obtain a more discriminative depth representation. Given a depth feature map $F^d \in \mathbb{R}^{H \times W \times C}$ and segmentation map $F^s \in \mathbb{R}^{H \times W \times C}$ from level $l$ with feature dimension equals $C$ and $W, H$ are width and height of feature maps on level l. The enhanced geometric representations can be generalized as:
\begin{equation}
\mbox{Fusion}\langle f_{i}, F^s \rangle =\sum_{j \in \Omega(F^s)} A_{i,j} \mathbf{W}_{i,j} f_{j},
\end{equation}
where $f_i \in F^d$ is query feature from depth image at position $i$, $A$ is assigned weight of $j_{th}$ feature point $f_j$ from sample space $\Omega(F^s)$ which is a subset of $F^s$ based on sampling function $\Omega$, and $\mathbf{W}_{i,j} \in \mathbb{R}^{C \times C}$ is the feature projection matrix. In \cite{guizilini2020semantically}, the distribution of the weights is calculated as the correlation between the guiding features and the Gaussian kernel. The projection matrix $\mathbf{W}$ are convolutional weights with kernel size $k$ and $\Omega$ is defined as a $k \times k$ convolutional window. \\
\indent{}Transformer attention is designed to capture global dependencies to build spatial interactions over enlarged areas. Specifically, let the sampling space $\Omega$ include all feature points from $F_s$. The attention function is performed on query vector $\mathbf{W}_q f^d$, key vector $\mathbf{W}_k f^s$ and value vector $\mathbf{W}_v f^s$, where $\mathbf{W}_q, \mathbf{W}_k$ and $\mathbf{W}_v$ are three separate linear transform functions. The attention weight is expressed as $A_{i,j} \propto \exp \left\{\frac{f_i^{T}\mathbf{W}_q^{T} \mathbf{W}_k f_j}{\sqrt{C}}\right\}$,
where $\sum_{j \in \Omega} A_{i j}=1$. Since the attention weights are distributed over the entire feature space, the cross-attention module suffers from slow convergence and computational complexity overhead on feature maps with large spatial sizes. An alternative way to cross-attention is deformable attention from \cite{zhu2020deformable}, which only attends to a small set of key sampling points around the query point with:
\begin{equation}
\mbox{Fusion}\left\langle f_{i}, F^{s}\right\rangle=\sum_{j \in \Omega(F^{s})} A_{i, j} \mathbf{W}_{i, j} f_{p_i + \Delta p_{i}}.
\end{equation}
In specific, the $\Omega$ samples $M$ pairs of key-value vectors from $F^s$. Those vectors are obtained from $F^s$ via interpolation on position $p_i + \Delta p_{i}$, where $p_i$ is the position coordinates of query point $f_i$; $\Delta p_{i} \in \mathbb{R} ^2$ denotes the sampling offset with unconstrained range, which are learned from linear functions. While the computational cost, especially on large spatial resolutions, can be significantly reduced using sampled key-value pairs, the sample space is still the same size as $F_s$. The experimental results show that the sample space is too large for the linear function to find the most relative points, yielding divergence attention and worse performance in high-resolution settings, as shown in Figure~\ref{fig:atten_compare}. \\
\begin{figure}[t]
\centering
\includegraphics[width=\linewidth]{ 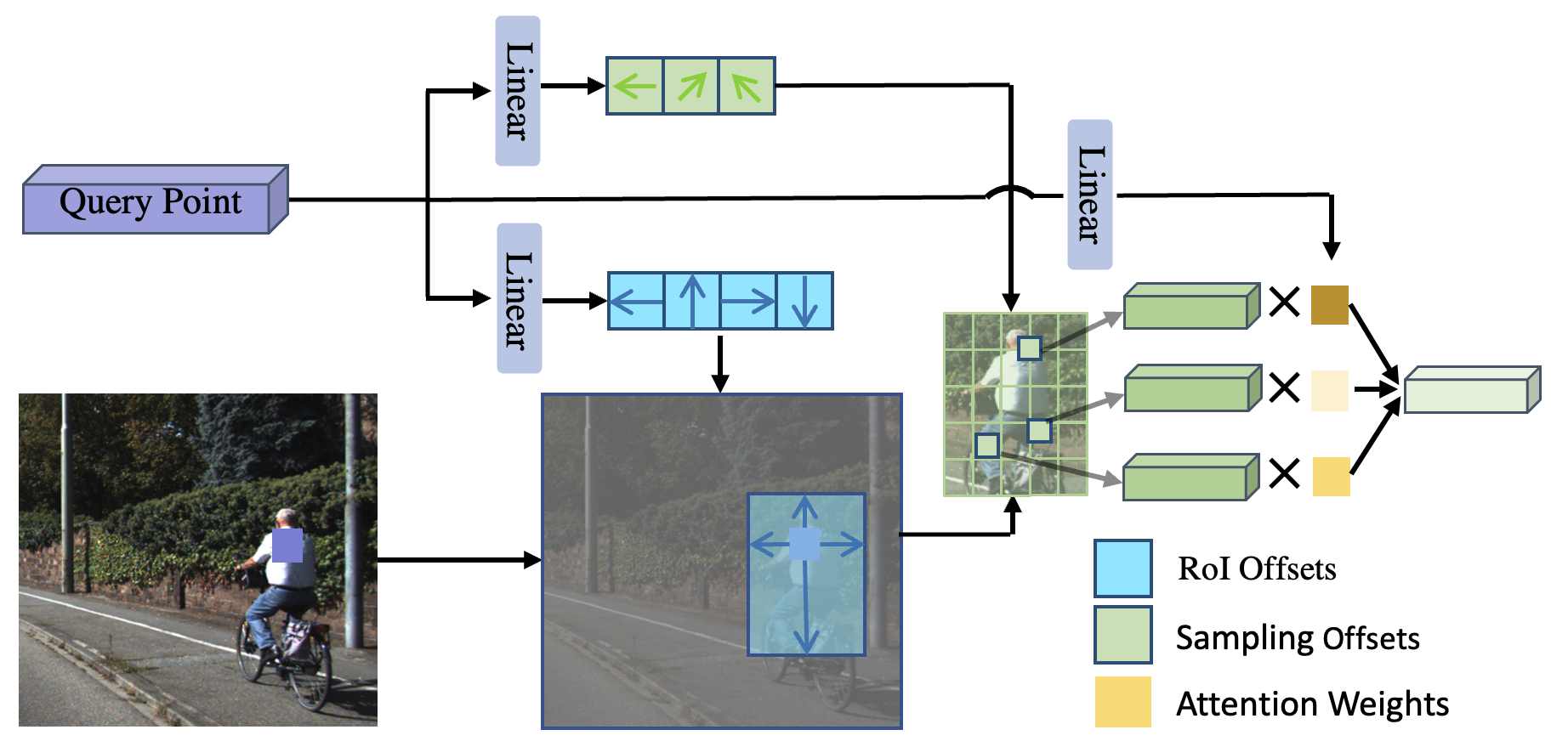}
\caption{ROI attention overview: a local region of interest is first inferred from semantic cues, using sampling offsets and attention weights with two separate linear functions.The features are sampled via interpolation without cropping the ROI region explicitly.}
\label{fig:atten1}
\end{figure}
%
\indent{}To improve the attention efficiency and facilitate the location of relative information in limited iterations, we constraint the attention in a local region which is much smaller than the original sample space. Inspired by \cite{yang2018metaanchor}, we generate the Region of Interests (ROIs) for each query point. Thanks to the supervision from segmentation, the local semantic aware ROIs can be inferred from $F^s$ efficiently. Specifically, we denote the ROI for a query point $f_i$ as bin $b_i = [d_l, d_t, d_r, d_b] \in [0,W]\times[0,H]\times[0,W]\times[0,H]$ which represents the distances from $f_i$ to the edges of the ROI box and $[w = d_l+ d_r,h = d_t + d_b]$ are normalized width and height.In practice, we constraint the width and height of ROIs with a maximum value, $r_{min}$ and $r_{max}$.Therefore, the feature fusion within a specific region of interest can be expressed as:
\begin{equation}
\mbox{Fusion}\left\langle f_{i}, b_i\right\rangle=\sum_{j \in \Omega(b_i)} A_{i, j} \mathbf{W}_{i, j} f_{p_i + \Delta \frac{1}{2} \cdot p_{i} \cdot b_i^{wh}},
\end{equation}
where the key-value pairs are sampled from $b_i$ via the interpolation on position $p_i + \Delta \frac{1}{2} \cdot p_{i} \cdot b_i^{wh}$, $ \Delta p_{i} \in \mathbb{R} ^2$ denotes the sampling offset with unconstrained range and $p_i$ is the center position of ROI. ROI size $b_i^{wh}$ guarantees the sampling offset is within $[-\frac{w}{2}, \frac{w}{2}] \times [-\frac{h}{2}, \frac{h}{2}]$. Without cropping and pooling explicitly, the sampling points can be obtained via interpolation over the neighboring  positions inside the ROIs. We simplify the ROI generation with linear function which outputs a bounding box based on the hypothesis that most relevant information are distributed around the query point. Moreover, the deformable attention mechanism enables sampling in any position within ROIs. The attention module is shown in Figure~\ref{fig:atten1}. By introducing a lightweight module, our semantic bounded attention module prompts the attention efficiency and concentrates computational resources via narrowing down the search space.
\subsection{ROI Former Module}
Inspired by \cite{carion2020enddetr}, the representative ability of ROI Attention can be boosted with a multi-head setting to capture guidance information from different semantic levels. Thus, we employ the multi-head structure to generate multiple proposals which provide separate ROIs for local attention. The multi-head ROI attention feature is:
\begin{equation} 
\hspace*{-3mm}f_i^d=\mbox{concat}(\mbox{Fusion}\left\langle f_{i}, b_i^1\right\rangle, \ldots, \mbox{Fusion}\left\langle f_{i},b_i^M\right\rangle) \mathbf{W}_O
\end{equation}
where $\mbox{Fusion}\left\langle f_{i}, b_i^m\right\rangle$ is the fusion feature after ROI attention on the $m_{th}$ head. The features from the $M$ head attention are merged by concatenation, following an output projection matrix $\mathbf{W}_O$. Beside the attention from depth to segmentation features, we apply the same multi-head ROI attention over depth feature to update semantic representation for mutual enhancement. To save the computational overhead when computing the guiding feature maps, we stack $f^s$ and $f^d$ at the beginning and use it as a shared guidance feature map after the convolutional layers, as shown in Figure~\ref{fig:framwork} . The shared guidance feature map is then fed into two attention blocks with $N$ stacked multi-head ROI attention layers to update depth and semantic feature separately. The outputs are fed into the upper level after upsampling to refine features on high resolution layers.  
\subsection{Network Architecture}
Our segmentation and depth estimation networks have the same architecture as in\cite{godard2019monodepth2}, i.e., a U-Net with skip connections, except for the depth fusion module. The feature maps of encoder $P={C6,C5,C4,C3,C2}$ from ~1/32 to 1/2 are extracted and fed into different levels of the encoder for dense prediction. Those feature maps are projected into $P={P6,P5,P4,P3,P2}$ of dimension $C={256, 128, 64, 32, 16}$ with five separate convolutional layers. The decoder consists of five upsample stages where, in each stage, the feature from the last decoding level is pre-processed with a convolutional layer and then upsampled and concatenated with features in the current level. The concatenated features are fed into another convolutional layer. The same operations are applied to the segmentation branch. Finally, the semantic features and depth features are fed into the ROIFormer module to obtain the fusion features,  as shown in Figure~\ref{fig:framwork}. Within ROIFormer, the depth feature and semantic feature are first concatenated into a common feature map. For efficiency purposes, we stack the feature maps from the segmentation branch and depth branch directly and use them as common attention memory. 
\subsection{Semantic Guided Re-projection Mask}
Similar to \cite{godard2019monodepth2}, we also predict depth maps $\hat{D}^{L}$ on intermediate layers to calculate projection loss on multiple scales. For the segmentation map, we only use the final predictions. Photometric loss calculation is less accurate  near the object boundaries and instance segments connected regions due to the depth estimation uncertainties. To overcome the boundary contamination problem, we create a mask for each point with a penalty coefficient according to the distance from the instance points to the areas. To this end, we build a graph from instance points set $\mathcal{S}_{Ins}$ to reference points set $\mathcal{S}_{Ref}$ and apply K-nearest-neighbors to sample the $K$ and get average relative distance $d_{t, i \rightarrow \mathcal{S}_{Ref}}$. Consequently, after re-projection into the consecutive images, we get the confidential mask as:
\begin{equation}
\mu_{t, i}=\left\{\begin{array}{cl}
\frac{1}{e^{\alpha d_{t,i}}}, & i \in \mathcal{S}_{ins}  \\
1,& i \notin \mathcal{S}_{ins}
\end{array}\right .
\end{equation}
\begin{figure}
\centering
\includegraphics[width=0.9\linewidth]{ 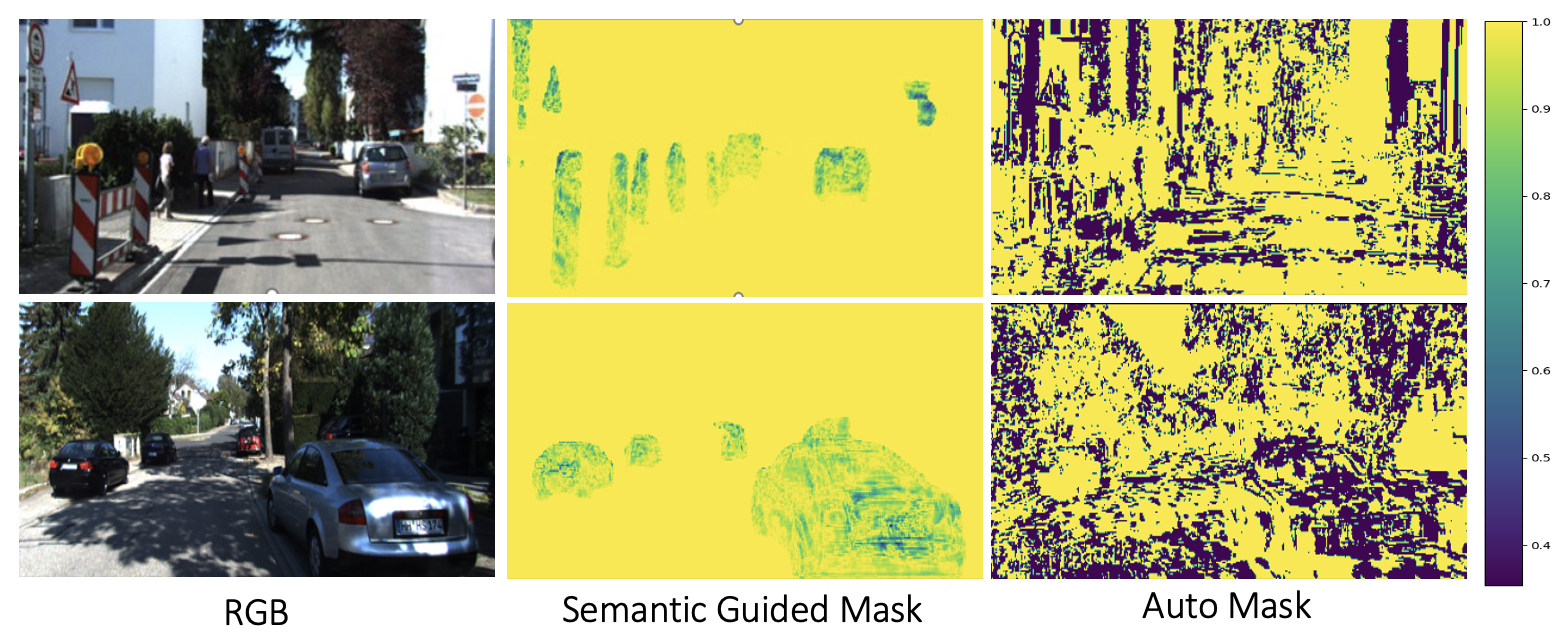}
\caption{Visualization comparison between the semantic guided re-projection mask (second column) and auto mask (third column). }
\label{fig:SBMask}
\end{figure}
We set 1 to each $i$th pixel position not belonging to instance points set $\mathcal{S}_{ins}$ and penalty weight to instance points obtained from an exponentially decreasing function with a scale factor $\alpha$. As shown in Figure~\ref{fig:SBMask}, compared with \cite{godard2019monodepth2}'s automask, our semantic bounded mask concentrate on the border areas where the depth value varies dramatically or reflective areas like windows of cars. In practice, we only consider the salient objects as reference points. The loss function of our model is obtained as :
\begin{equation}
L = \mu * L_{p}+\beta \cdot L_{s} + \gamma \cdot L_{sem}.
\end{equation}
where $\beta$ and $\gamma$ are weighted factors for smoothness loss and segmentation loss.
\section{Experimental Results}
\label{sec:results}


\subsection{Implementation Details}
Our encoder is built using Resnet-18 and Resnet-50 backbones pre-trained on the ImageNet. The pose network is a pre-trained Resnet-18 model. The input image size is set to a medium resolution of $192 \times 640$ and a high resolution of $320 \times 1024$ following the same setting in the former methods for a fair comparison. We train our model with a batch size of 12 on a single NVIDIA Tesla V100 GPU. Following \cite{godard2019monodepth2}, all experiments are trained with 20 epochs with the learning rate of $10^{-4}$ and decayed by 10 on the 10th and 15th epochs. All experiments are trained with an ADAM optimizer.
For supervised segmentation training, we employ the Cross-Entropy loss with a hyper-parameter $\gamma = 0.5$ to control the weights of segmentation loss. The $\alpha$ for photometric loss is set to 0.85 and $\beta = 1 \times 10^{-3}$. The ROIs range is set to $r_{min}=\{0.3W,0.3H\}$ and $r_{max} = \{0.7W, 0.7H\}$  For evaluation, the input image size is the same as the training set, and the final output of the model is resized into the ground truth resolution ($384 \times 1280$). We apply the ground truth scale technique to restore the absolute depth estimation, and the depth values are restricted to the range of 0 to 80 meters. No further post-processing is required.\\
\indent{}The KITTI dataset is used for self-supervised depth evaluation using the same settings as in~\cite{zhou2017unsupervised} to remove the static frames from the data split of~\cite{eigen2015predicting}, resulting in 39810 images used for training, 4424 for validation ,and 697 for final evaluation. 
\subsection{Ablation Study}

\begin{table}[htbp]
\resizebox{\columnwidth}{!}{%
\begin{tabular}{ccccc}
\hline
$\#$ of attention layers & AbsRel & SqRel & RMSE & RMSElog \\ \hline
0&1.037&0.692&4.428&0.179\\
1&1.011&0.679&4.323&0.175\\
2&1.002&0.667&4.320&0.174\\
3&1.015&0.617&4.359&0.176\\
4&1.027&0.678&4.364&0.177\\ \hline
\end{tabular}%
}
\caption{Ablation study on the number of attention layers. The number of attention layers are shared across P3 to P5.}
\label{tab:attenlayers}
\end{table}
\subsubsection{ROIFormer Module}
ROIFormer is the core operation in the proposed method, which guarantees the efficient mutual information fusion between semantics and depth and the feature representation enhancement. We first investigate the impact of the number of attention layers, where the attention block is applied on P3, P4, and P5 levels with $N$ attention layers stacked for both segmentation and depth queries. Thanks to the shared memory and local attention design, the memory cost is linear to the input feature size, making it possible to test more stacked layers. As shown in Table~\ref{tab:attenlayers}, removing the attention blocks degenerate the fusion layer into a simple concatenation operation. The interaction between two domains relies on stacked convolutional operations, which provides limited feature enhancement. Adding one attention layer yields significant improvement over all metrics. We found that using two stack layers achieves the best trade-off between accuracy and complexity while additional layers contribute less or even hinder the performance. \\
\begin{table}[htbp]
\resizebox{\columnwidth}{!}{%
\begin{tabular}{cccc|cccc}
\hline
P5 & P4 & P3 & P2 & AbsRel & SqRel & RMSE & RMSElog \\ \hline
\checkmark&&&&0.103&0.7049&4.437&0.1785\\
\checkmark&\checkmark&&&0.1008&0.7004&4.392&0.1774\\
\checkmark&\checkmark&\checkmark&&0.1003&0.6844&4.366&0.1764\\
\checkmark&\checkmark&\checkmark&\checkmark&0.1002&0.6603&4.330&0.1756\\ \hline
\end{tabular}%
}
\caption{Ablation study on the impact of layers for feature fusion. P2-inclusion improves the performance, while P3-P5 was only used in all other experiments.}
\label{tab:fusionlayers}
\end{table}
\indent{}We further investigate the impact of attention layers on different pyramid levels and show results in Table~\ref{tab:fusionlayers}. The experiments show that feature fusion only on deep layers with limited resolution leads to a minor improvement in depth precision. Shallow layers with fewer feature channels bring more fine-grained details, resulting in compelling results. We further test our attention module on P2 level with 1/2 input size resulting in similar results. Finally, we compare the two sampling strategies and the effects of attention points in the attention module. Similar to the aforementioned cases, the features in shallows are vital for final outputs. Thus, it is reasonable that assigning more attention points in shallow layers brings more performance benefits. In our experiments, the combination of [8,16,32] is the best setting considering the precision and efficiency trade-off. 
\begin{table}[htbp]
\resizebox{\columnwidth}{!}{%
\begin{tabular}{c|c|cccc}
\hline
Type                             & \# of att points & AbsRel & SqRel & RMSE & RMSElog \\ \hline
\multirow{3}{*}{Fixed}           & {[}8{]}          &0.1036&0.7269&4.416&0.1777\\
                                 & {[}16{]}         &0.1009&0.7027&4.395&0.1770\\
                                 & {[}32{]}         &0.1012&0.7164&4.414&0.1773\\ \hline
\multirow{3}{*}{Total}           & {[}4,8,16{]}     &0.1010&0.6972&4.386&0.1754\\         
                                 & {[}8,16,32{]}    &0.1003&0.6855&4.360&0.1745\\
                                 & {[}16,32,64{]}   &0.0998&0.7137&4.415&0.1771\\ \hline
\end{tabular}%
}
\caption{Ablation study on attention points sampling strategies and number of points. The features of pyramid layers are fixed, as well as the attention head numbers.}
\label{tab:attenpoints}
\end{table}

\begin{table}[htbp]
\centering
\resizebox{\columnwidth}{!}{%
\begin{tabular}{c|cc|cc|c}
\hline
Attention Type & Resolution & Net & Sq Rel & RMSE & $\theta < 1.25$ \\ \hline
Transformer&640&Res18&0.722&4.547&0.886\\
Transformer&1024&Res18&0.687&4.366&0.895\\
Transformer&640&Res50&0.675&4.393&0.893\\
Transformer&1024&Res50&$\times$&$\times$&$\times$\\ \hline
Deformable Attn&640&Res18&0.717&4.472&0.8871\\
Deformable Attn&1024&Res18&0.7042&4.391&0.8795\\
Deformable Attn&640&Res50&0.6961&4.399&0.8942\\
Deformable Attn&1024&Res50&0.7383&4.517&0.8696\\ \hline
ROIFormer&640&Res18&0.6959&4.438&0.8908\\
ROIFormer&1024&Res18&0.6749&4.335&0.897\\
ROIFormer&640&Res50&0.6733&4.351&0.895\\
ROIFormer&1024&Res50&0.6161&4.148&0.9041\\ \hline
\end{tabular}%
}
\caption{Comparison between different attention types}
\label{tab:atten_compare}
\end{table}

\begin{figure}
\centering
\includegraphics[width=0.84\linewidth, height=115pt]{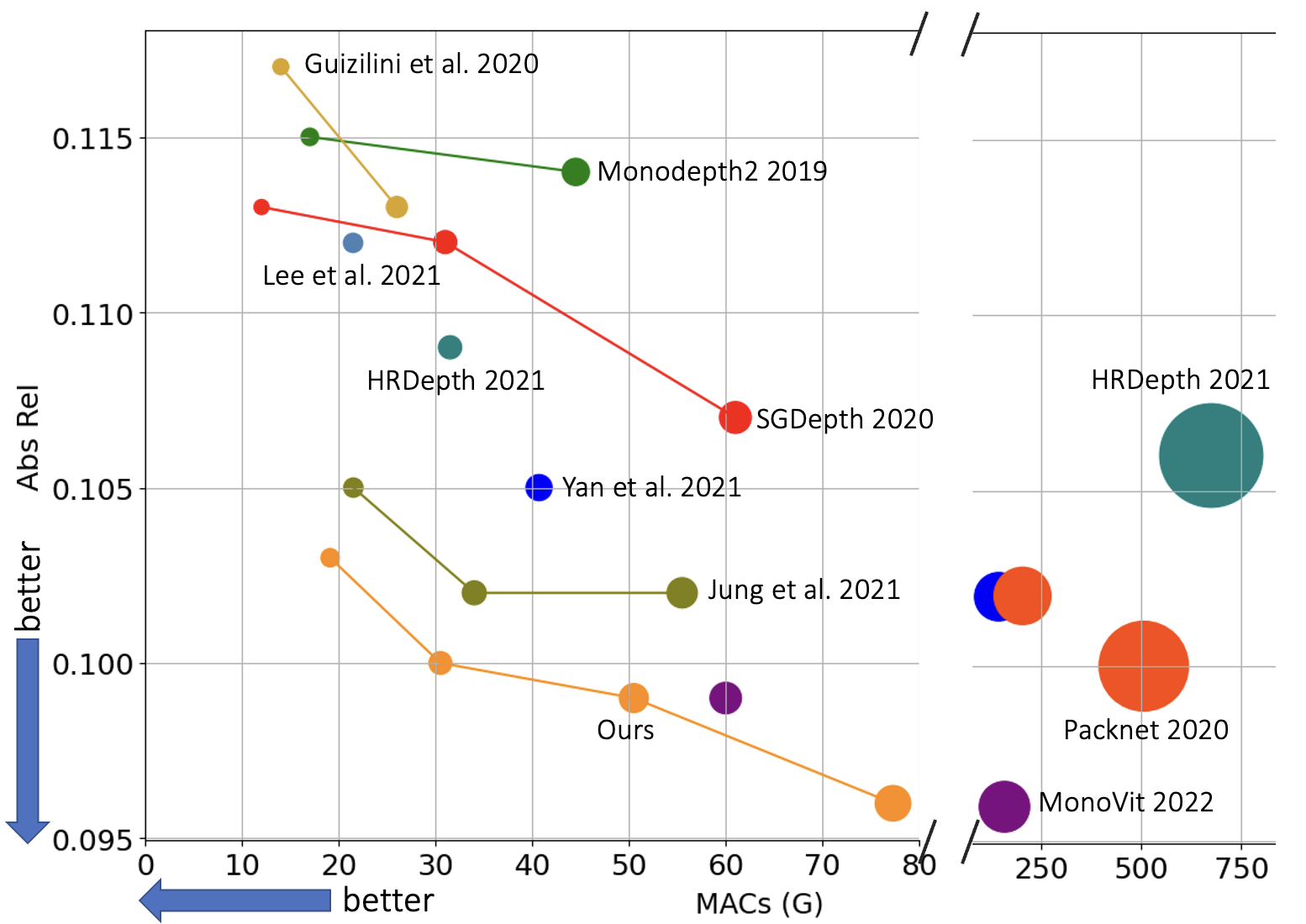}
\caption{Speed-Accuracy trade-off curve.}
\label{fig:complexity}
\end{figure}

\subsubsection{Complexity Comparison}
\begin{figure}
\centering
\includegraphics[width=0.48\textwidth]{ 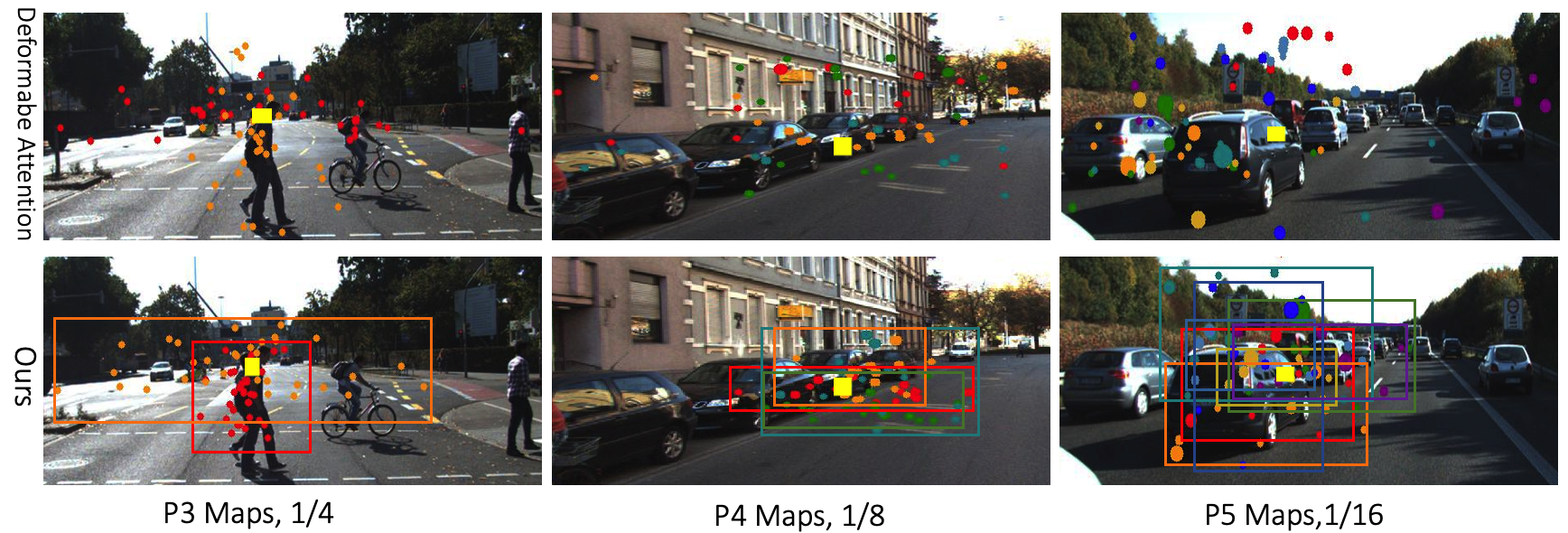}
\caption{Attention visualization of ROIFormer and Deformable Attention on multi-scale heads in different colors. The query points are marked as yellow rectangles. Our ROIFormer learned most relevant references while deformable attention diverged.}
\label{fig:attention_vis}
\end{figure}

The model efficiency of our proposed method is explored by comparing it with other attention-based feature fusion methods, i.e., transformer attention and deformable attention. As shown in Table\ref{fig:atten_compare}, the transformer performs better than deformable attention but exceeds the memory limit on HR inputs. The performance of deformable attention on HR inputs drops dramatically. We argue that the sample space is too large to select valid key-value pairs, resulting in divergence. ROIFormer, achieves the best performance across all settings, demonstrating its superiority over other attention methods. Figure.~\ref{fig:attention_vis} shows the attention areas and sampled key values based on different attention variants. ROIFormer samples the relative features more efficiently. We further summarize the model complexity against the performance of all STOA methods and plot the trade-off curve in Figure.\ref{fig:complexity}. The model size is shown as areas of circles. 
\begin{table}[]
\resizebox{\columnwidth}{!}{%
\begin{tabular}{c|c|c|c|cccc|c}
\hline
M & Sem & ROIFormer & Mask & AbsRel & SqRel & RMSE & RMSElog & $\delta < 1.25$ \\ \hline
\checkmark&&&&0.115&0.903&4.863&0.192&0.877\\
\checkmark&&\checkmark&&0.108&0.79&4.595&0.184&0.888\\
\checkmark&\checkmark&\checkmark&&0.1005&0.6733&4.351&0.1756&0.895\\
\checkmark&\checkmark&\checkmark&\checkmark&0.1002&0.654&4.356&0.175&0.898\\ \hline
\end{tabular}%
}
\caption{Ablation study for the contribution of three main components: monocular (M) training, semantics (Sem) information, ROIFormer and semantics guided mask loss. }
\label{tab:ablation1}
\end{table}
\begin{table*}[htbp]
\resizebox{\textwidth}{!}{%
\begin{tabular}{c|c|c|c|cccc|ccc}\hline
Method & Backbone  & Sem & Resolution & AbsRel & SqRel & RMSE & RMSElog & $\theta < 1.25$ & $\theta < 1.25^2$ & $\theta < 1.25^3$ \\ \hline
\cite{zou2020learning}&Resnet-18&&$192\times640$&0.115&0.871&4.778&0.191&0.874&0.963&\textbf{0.984}\\
\cite{johnston2020self}&Resnet-18&&$192\times640$&0.111&0.941&4.817&0.185&0.885&0.961&0.981\\ 
\cite{hui2022rm}&Resnet-18&&$192\times640$&0.108&0.71&4.513&0.183&0.884&0.964&0.983\\
\cite{jung2021fineFSRE}&Resnet-18&\checkmark&$192\times640$&0.105&0.722&4.547&0.182&0.886&0.964&\textbf{0.984}\\
\textbf{Ours}&Resnet-18&\checkmark&$192\times640$&\textbf{0.103}&\textbf{0.6959}&\textbf{4.438}&\textbf{0.1778}&\textbf{0.8892}&\textbf{0.9648}&0.9836\\ \hline
\cite{klingner2020sgdepth}&Resnet-18&\checkmark&$384\times1280$&0.107&0.768&4.468&0.186&0.891&0.963&0.982\\
\cite{choi2020safenet}&Resnet-18&&$320\times1024$&0.106&0.743&4.489&0.181&0.884&0.965&\textbf{0.984}\\
\cite{lyu2021hrdepth}&Resnet-18&&$320\times1024$&0.106&0.755&4.472&0.181&0.892&0.966&\textbf{0.984}\\
\cite{jung2021fineFSRE}&Resnet-18&\checkmark&$320\times1024$&0.102&0.687&4.366&0.178&0.895&\textbf{0.967}&\textbf{0.984}\\
\textbf{Ours}&Resnet-18&\checkmark&$320\times1024$&\textbf{0.100}&\textbf{0.6749}&\textbf{4.335}&\textbf{0.1757}&\textbf{0.8962}&0.9665&0.9836\\ \hline
\cite{godard2019monodepth2}&Resnet-50&&$192\times640$&0.115&0.903&4.863&0.193&0.877&0.959&0.981\\
\cite{guizilini2020semantically}&Resnet-50&\checkmark&$192\times640$&0.113&0.831&4.663&0.189&0.878&0.971&0.983\\
\cite{kumar2021syndistnet}&Resnet-50&\checkmark&$192\times640$&0.109&0.718&4.516&0.18&\textbf{0.896}&\textbf{0.973}&\textbf{0.986}\\
\cite{yan2021channel}&Resnet-50&&$192\times640$&0.105&0.769&4.535&0.181&0.892&0.964&0.983\\
\cite{li2021learning}&Resnet-50&\checkmark&$192\times640$&0.103&0.709&4.471&0.18&0.892&0.966&0.984\\
\cite{jung2021fineFSRE}&Resnet-50&\checkmark&$192\times640$&0.102&0.675&4.393&0.178&0.893&0.966&0.984\\
\textbf{Ours}&Resnet-50&\checkmark&$192\times640$&\textbf{0.100}&\textbf{0.6733}&\textbf{4.351}&\textbf{0.1756}&0.8958&0.9665&0.9848\\ \hline
\cite{godard2019monodepth2}&Resnet-50&&$320\times1024$&0.115&0.882&4.701&0.19&0.879&0.961&0.982\\
\cite{shu2020feature}&Resnet-50&&$320\times1024$&0.104&0.729&4.481&0.179&0.893&0.965&0.984\\
\cite{gurram2021monocular}&Resnet-50&&$320\times1024$&0.104&0.721&4.396&0.185&0.88&0.962&0.983\\
\cite{kumar2021syndistnet}&Resnet-50&\checkmark&$320\times1024$&0.102&0.701&4.347&\textbf{0.166}&0.901&0.98&\textbf{0.99}\\
\cite{cai2021xdistill}&Resnet-50&\checkmark&$320\times1024$&0.102&0.698&4.439&0.18&0.895&0.965&0.983\\
\cite{chanduri2021camlessmonodepth}&Resnet-50&&$320\times1024$&0.102&0.723&4.374&0.178&0.898&0.966&0.983\\
\cite{petrovai2022exploiting}&Resnet-50&&$320\times1024$&0.101&0.72&4.339&0.176&0.898&0.967&0.984\\
\textbf{Ours}&Resnet-50&\checkmark&$320\times1024$&\textbf{0.096}&\textbf{0.6161}&\textbf{4.148}&0.1697&\textbf{0.9045}&\textbf{0.9692}&0.9856\\ \hline
\end{tabular} 
}
\caption{Comparison with the state-of-the-art on KITTI Eigen test set.}
\label{tab:results}
\end{table*}
\subsubsection{Impact of Individual Component.} We summarize the impact on depth estimation accuracy of each main component in Table~\ref{tab:ablation1}. For the baseline, we keep only self-supervised depth estimation modules including smoothness loss and photometric loss. Applying our attention module only to the intermediate depth features brings over $6\%$ improvement on relative errors which imply the significant impact of attention on feature representation enhancement. Together with semantics information, the attention module is capable of getting the most of mutual benefits from segmentation and depth domains.
\begin{figure}
\centering
\includegraphics[width=0.48\textwidth]{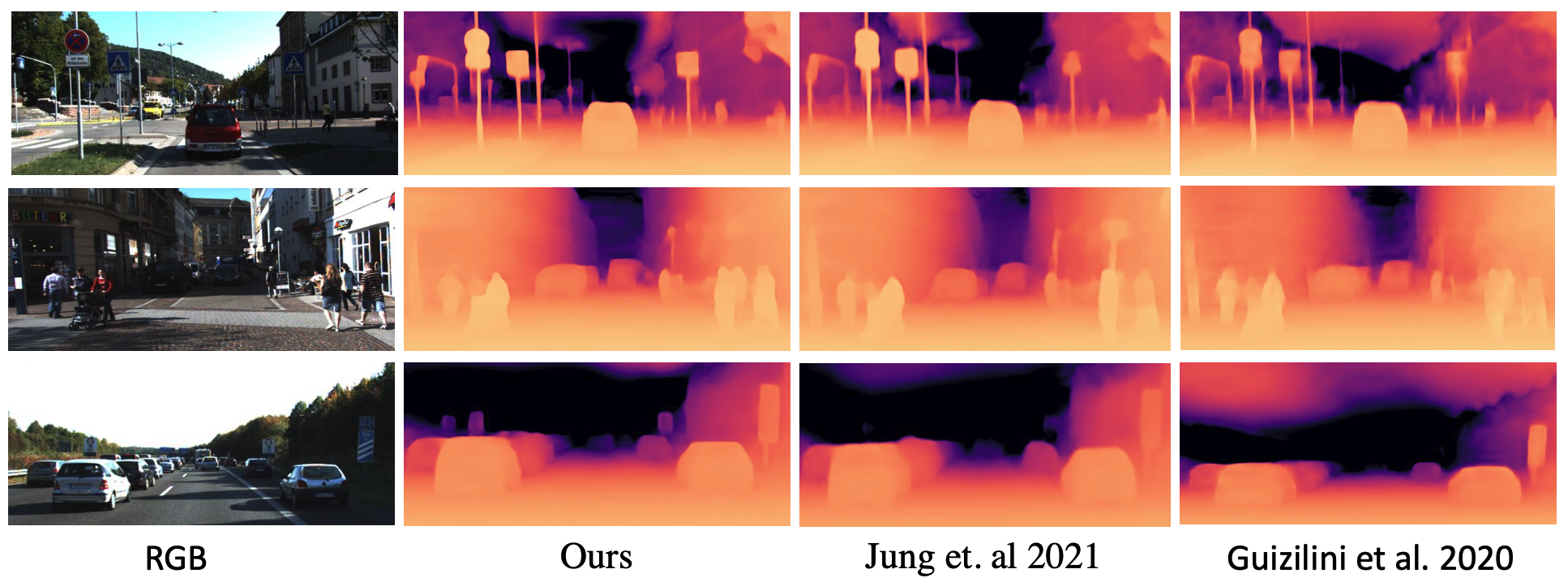}
\caption{Qualitative self-supervised monocular depth estimation performance comparing ROIFormer with previous State-of-the-Art.}
\label{fig:results}
\end{figure}
\subsection{Comparison with State-Of-The-Art Methods}
We evaluate ROIFormer on the KITTI dataset based on the metrics from \cite{eigen2015predicting}. As shown in Table~\ref{tab:results}, our proposed method outperforms all existing SOTA self-supervised monocular depth estimation methods, including approaches utilizing semantic information. ROIFormer enables flexible local feature interaction, resulting in a much higher performance under HR settings with more fine-grained details. We also compare the model complexity by calculating the MACs and model parameter numbers. The trade-off curve is shown in Figure~\ref{fig:complexity}. We achieve the same performance with only $30\%$ MACs, compared to the recent SOTA method \cite{zhao2022monovit} which utilizes the transformer as the backbone. With Resnet18, the ROIFormer runs at 51 fps on MR and 33 fps on HR. With Resnet50, it runs at 45 and 24 fps respectively. Finally, Figure~\ref{fig:results} illustrates the performance comparison with other SOTA methods qualitatively.
\section{Conclusion}
In this paper, we propose a novel attention module to improve the self-supervised depth estimation accuracy. Our method enhances the representative feature ability by learning spatial dependencies from local semantic areas. We leverage geometric cues from segmentation feature maps to learn the regions of interest to guide feature aggregation cross domains. The attention module is employed in a multi-head and multi-scale schema to enable the feature learning from different semantic levels. Furthermore, we introduce a semantic aware projection mask to improve the model robustness on uncertain areas.We conducted extensive experiments on the KITTI dataset and achieved new SOTA performance, which demonstrated its effectiveness.

\bibliography{aaai23}

\end{document}